\documentclass{article}
\usepackage{ijcai26}

\usepackage{comment}
\usepackage{multirow}
\usepackage{times}
\usepackage{soul}
\usepackage{url}
\usepackage[hidelinks]{hyperref}
\usepackage[utf8]{inputenc}
\usepackage[small]{caption}
\usepackage{graphicx}
\usepackage{amsmath}
\usepackage{amsthm}
\usepackage{booktabs}
\usepackage{algorithm}
\usepackage{algorithmic}
\usepackage{float}
\usepackage[switch]{lineno}
\usepackage{amssymb}

\title{Unified Multi-Dataset Training for TBPS}

\author{
Nilanjana Chatterjee$^{\dagger,1}$
\and
Sidharatha Garg$^{\dagger,1}$\and
A V Subramanyam$^{1}$\And
Brejesh Lall$^2$\\
\affiliations
$^1$IIIT Delhi, $^2$IIT Delhi\\
\emails
nilanjanac@iiitd.ac.in, sidhartha22499@iiitd.ac.in, subramanyam@iiitd.ac.in, brejesh@ee.iitd.ac.in
}

\begin{document}

\maketitle
\begingroup
\renewcommand\thefootnote{}
\footnotetext{$\dagger$ denotes equal contribution}
\endgroup

\begin{abstract}

    Text-Based Person Search (TBPS) has seen significant progress with vision–language models (VLMs), yet it remains constrained by limited training data and the fact that VLMs are not inherently pre-trained for pedestrian-centric recognition. Existing TBPS methods therefore rely on dataset-centric fine-tuning to handle distribution shift, resulting in multiple independently trained models for different datasets. While synthetic data can increase the scale needed to fine-tune VLMs, it does not eliminate dataset-specific adaptation. This motivates a fundamental question: can we train a single unified TBPS model across multiple datasets? We show that naive joint training over all datasets remains sub-optimal because current training paradigms do not scale to a large number of unique person identities and are vulnerable to noisy image–text pairs. To address these challenges, we propose Scale-TBPS with two contributions: (i) a noise-aware unified dataset curation strategy that cohesively merges diverse TBPS datasets; and (ii) a scalable discriminative identity learning framework that remains effective under a large number of unique identities. Extensive experiments on CUHK-PEDES, ICFG-PEDES, RSTPReid, IIITD-20K, and UFine6926 demonstrate that a single Scale-TBPS model outperforms dataset-centric optimized models and naive joint training. Code is available.
\end{abstract}

\section{Introduction}
Text-based Person Search (TBPS) has emerged as a critical task in computer vision, bridging the gap between natural language understanding and fine-grained visual retrieval to locate specific individuals in unconstrained image galleries. Since its formal introduction by \cite{b13} alongside the benchmark CUHK-PEDES dataset, the field has expanded through numerous benchmarks in recent years including ICFG-PEDES \cite{b6}, RSTPReid \cite{b5}, IIITD-20K \cite{b14}, and UFine6926 \cite{b41}. The inception of Vision-Language Models (VLMs) such as CLIP \cite{b12} and ALBEF \cite{b21} has yielded significant performance gains. However, these models are primarily trained on broad online corpora \cite{b42,b43}. Though they show notable zero shot results in multiple domains, the performance in zero shot fine-grained domains still remains unsatisfactory \cite{b45}. 
To tackle this issue, various TBPS \cite{b16,b21,b23,b1} resort to dataset level fine-tuning and  propose several loss formulations and local feature networks to improve fine-grained feature extraction. Despite these advancements, the current paradigm remains heavily dataset-centric. Fine-tuning models on isolated benchmarks leads to poor cross-dataset generalization, as these models fail to account for the diverse distribution shifts caused by varying geographical locations, lighting conditions, and environmental contexts. This highlights the need for a unified approach to scale models across different shifts to maintain one unified model for all shifts.
\begin{figure}[t]
    \centering
    \includegraphics[width=\linewidth]{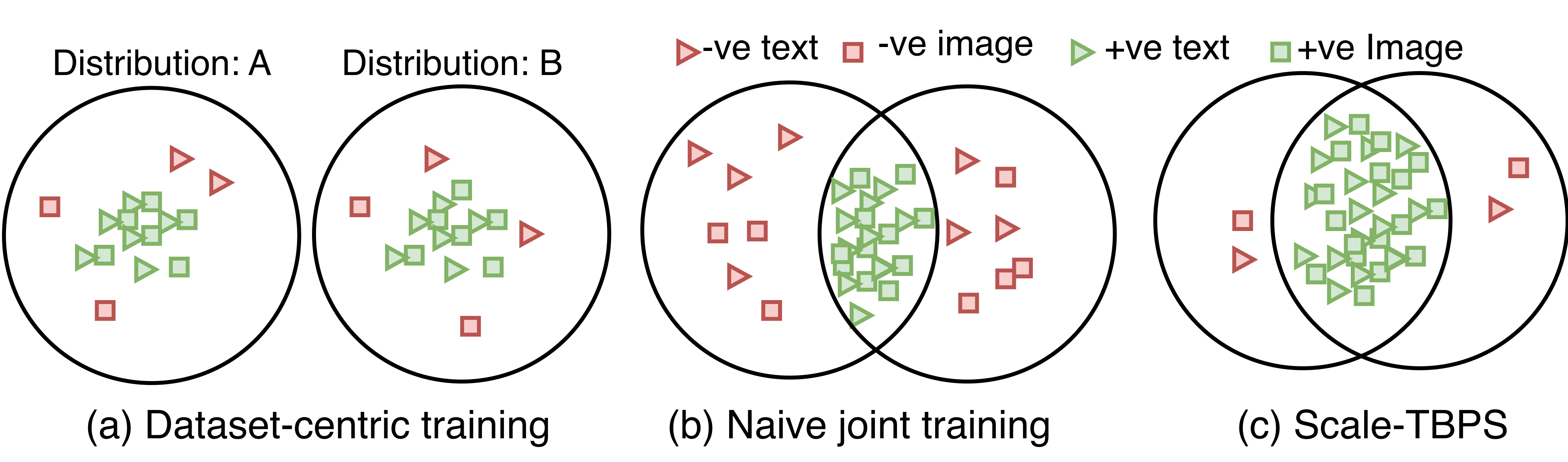}
    \caption{Illustration of Scale-TBPS. (a) illustrates the conventional dataset-centric training paradigm, where separate models are independently trained for different distributions, resulting in isolated models.
(b) depicts naive joint training, where a single model is trained on merged datasets; however the learned representation fails to adequately accommodate samples from all distributions.
(c) presents our proposed method, in which a single unified model is trained cohesively across distributions, effectively capturing shared semantics}
    \label{fig:abstract}
\end{figure}

While recent literature has explored the necessity of unified training in other vision tasks \cite{b44}, dedicated effort towards scaling in TBPS requires deeper investigation. In Figure \ref{fig:abstract}, we show an illustrative example of how dataset-oriented models remain limited to their own specific distributions. Even when naive joint training brings these distributions into a common space, it yields suboptimal scaling performance for TBPS, despite having access to more training data. This occurs because the distribution shift originating from multiple benchmarks and the large number of unique person IDs are not explicitly modeled in standard TBPS paradigms. Some recent works \cite{b22,b51} explore person-specific pretraining from scratch; however, they still require dataset-specific fine-tuning to achieve competitive performance. Moreover, individual TBPS datasets are relatively small compared to the large-scale data typically required for training vision–language models. A model trained on unified and curated data from multiple sources can leverage greater data diversity and scale, thereby achieving superior retrieval performance.

In this work, we explore the challenges of joint training and design Scale-TBPS with two core components: a dataset curation framework to create a cohesive, noise-filtered unified dataset, and a training strategy to address the large scale of unique IDs that arises out of the unification of datasets. While conventional TBPS training paradigms are not well-suited to scale across large and diverse datasets, our approach is specifically designed to accommodate the merging of multiple distribution shifts in a scalable manner. Our contributions are as follows.

\textbf{Noise-Aware Unified Dataset Curation (NDC).} We first study joint training and find that naively combining multiple datasets disregards noise and results in a model that does not leverage the large scale of the data. 
RDE \cite{b16} highlights caption noise in TBPS datasets but its iterative filtration process is not optimal for training with large-scale data.  Similarly another noise filtration work, WORA-TBPS \cite{b39} use unaligned models for evaluating the samples. 

A recent work advocates combining multiple signals or metrics to improve bias or limitations \cite{b57}. Therefore, our proposed NDC considers a set of diverse expert models trained on varied TBPS distributions. We find that some true positive pairs are far apart to be recognized, even by optimized TBPS models. Therefore, NDC relies on relative similarity ranking between text–image pairs rather than adopting a hard threshold. 
Combining consensus from diverse expert models enables more cohesive curation of the unified dataset. Our method is a one-time preprocessing step, making it effective for large-scale data.

\textbf{Discriminative ID Learning (DIL).} We introduce a training objective specifically to stabilize identity learning across diverse distribution shifts, ensuring discriminative feature embeddings. Existing TBPS methods incorporate identity-discriminative losses; however, under joint training across multiple datasets, we observe that such dataset-centric approaches consistently underperform their individually trained counterparts, despite being trained on a larger data pool. This indicates that conventional classifier-based losses struggle when the number of identities grows substantially; this phenomenon is also observed in large-scale classification settings such as face recognition \cite{b54}. In the face recognition domain, angular margin–based losses have demonstrated strong discriminative capability \cite{b40}. 

However, these formulations are primarily designed for unimodal vision tasks and are not directly applicable to text-based person search. To this end, we design a Multimodal Angular Identity loss for TBPS, which performs well with more data, overcoming the bottleneck of ID explosion of standard TBPS methods. As illustrated in Figure \ref{fig:abstract}, our method brings more scalability.

\section{Related Works}

\textbf{Text-based Person Search.} TBPS aims to identify the best-matched person image given a textual description and can be regarded as a fine-grained subtask of text-based image retrieval. Owing to its practical relevance and fine-grained nature, TBPS has attracted sustained research interest. Early works primarily relied on unimodal or weakly aligned frameworks, where bridging the modality gap between visual and textual representations received a great interest \cite{b8,b10}.
\begin{figure*}[t]
    \centering    \includegraphics[width=\textwidth]{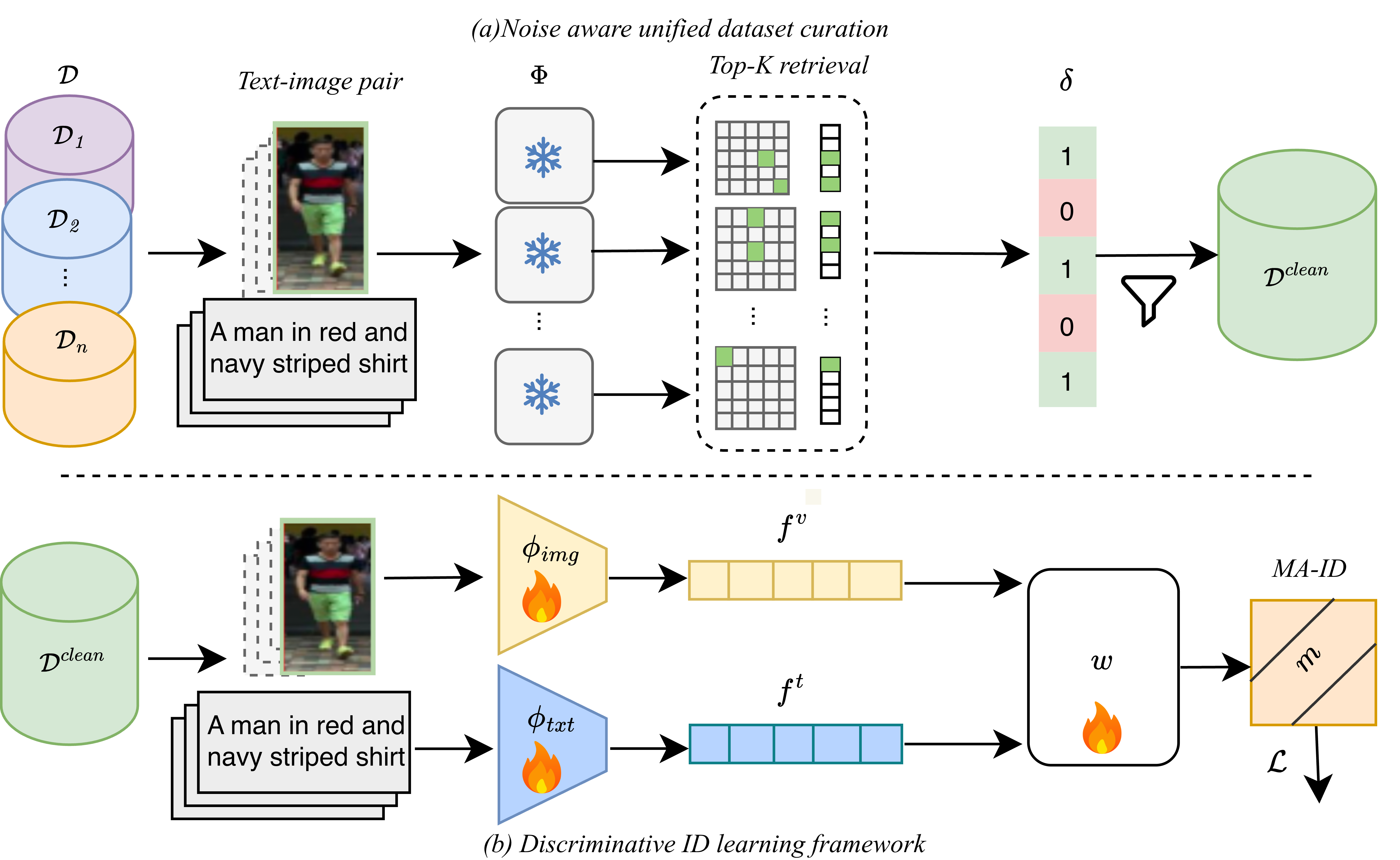}
    \caption{{Overview of the proposed Scale-TBPS.}
(a) {Noise-Aware Data Curation (NDC)}:
Text--image pairs from the joint dataset ($\mathcal{D}$) are encoded using a set of pretrained and frozen models $\Phi$.
top-$K$ retrieved samples are computed independently for each model. A pair is retained as a clean sample if it is ranked within the top-$K$ results by at least one pretrained model; such selected pairs are highlighted in green.
(b) {Discriminative Identity Learning (DIL)}: 
Filtered text--image pairs from $\mathcal{D}^{\text{clean}}$ are encoded by
the learnable image and text encoders and encoded features $f^{v}$ and $f^{t}$ are passed through a shared multimodal class weight $w$.
MA-ID denotes the proposed multimodal angular identity loss while $m$ denotes the margin}
\label{fig:method}
\end{figure*}

With the advent of large-scale VLMs, TBPS methods have significantly benefited from improved cross-modal alignment. In particular, CLIP has become a popular choice in TBPS due to its lightweight architecture and strong transferability, and has been widely adapted for this task \cite{b1,b2,b12,b23,b16}. ALBEF, although computationally heavier, provides stronger alignment capabilities and has been adopted in methods such as RaSa \cite{b24}.

More recently, another research direction has emerged that focuses on TBPS-specific pre-trained backbones \cite{b48,b51}, where large multimodal language models (MLLMs) are used to generate dataset-level captions for pre-training. While these approaches improve representation learning, they still exhibit limited zero-shot performance and require subsequent dataset-specific fine-tuning to achieve competitive results. Consequently, even after TBPS-specific pre-training, the reliance on individual dataset-specific models persists. This observation motivates the need to investigate whether TBPS can be scaled towards a single unified model that can handle diverse distribution shifts.

\textbf{Unified Paradigm.} Although VLMs demonstrate promising zero-shot capabilities, fine-grained tasks such as TBPS and vision-based person re-identification still require task-specific fine-tuning to achieve reliable performance. Recent research trends increasingly explore unified paradigms that aim to support multiple in-domain distribution shifts within a single model. For example, ATReID \cite{b47} addresses illumination variations by enabling person identification across different times of day, while UniHCP \cite{b44} trains a unified model for multiple human-centric perception tasks. Incremental learning has also been explored as a strategy to continually update models without catastrophic forgetting \cite{b53,b38}.

TBPS datasets exhibit substantial variation and it is impractical to assume a fixed, closed-world dataset. As new data continuously emerge due to environmental and domain variations, TBPS naturally calls for scalable and unified training strategies. In this work, we propose a dataset curation strategy that enables unified training  and a discriminative ID learning framework to handle a large number of unique identities.

\section{Methodology}

  Scale-TBPS aims to learn a single unified model that can handle data scaling across all available TBPS datasets. Our method explicitly targets the cohesive merging of different data distributions while handling large volumes of unique person identities. We formulate our problem in Section 3.1, followed by our noise-aware unified dataset curation strategy (NDC) in Section 3.2. Section~3.3 explains our Discriminative Identity Learning framework (DIL) designed to scale effectively with the number of person identities.

\subsection{Problem Formulation}
Let there be $n$ train-sets $\{\mathcal{D}_1, \mathcal{D}_2, \ldots, \mathcal{D}_n\}$, each containing image--text pairs annotated with person identities. These datasets are merged to form a joint dataset:
\begin{equation}
\mathcal{D} = \bigcup_{j=1}^{n} \mathcal{D}_j = \{I_i, T_i, y_i\}_{i=1}^{N},
\end{equation}
where $I_i$ and $T_i$ denote the $i$-th pedestrian image and its corresponding textual description, respectively, and $y_i$ represents a unique person identity across the unified dataset. We employ image encoder $\phi_{img}$ and text encoder $\phi_{txt}$ initialized with CLIP weights. These encoders are jointly optimized during training to produce aligned global visual and textual representations.


\textbf{Baseline.}
We adopt established dataset-centric TBPS model architectures as our representative baselines. To enable scalability, we design a training pipeline that can be applied on top of VLM baseline models, without modifying their core architectures.

\textbf{Feature Extraction.}
For each image $I_i$, we extract a global visual representation $f^{v}_i$ using an image encoder, where ${f}^v_{i}$ corresponds to the \texttt{[CLS]} token or an equivalent global pooled feature. Similarly, for each text description $T_i$, a global textual representation ${f}^t_{i}$ is obtained from a text encoder, using the \texttt{[EOS]} token.


\subsection{Noise-Aware Unified Dataset Curation (NDC)}

Due to annotation noise and cross-dataset distribution shifts, not all image-text pairs exhibit reliable semantic correspondence. Learning alignment confidence directly from untrained models is unreliable. Therefore, we adopt a noise-aware filtering strategy using a set of pretrained TBPS models. We perform the filtering only once and use the filtered dataset for training.

We consider a set of $p$ pretrained person-specific models
$\Phi = \{\phi_1, \phi_2, \ldots, \phi_p\}$, where $p < n$. This design enables the use
of diverse pretrained experts while maintaining scalability, since additional
datasets can be incorporated even when the corresponding pretrained TBPS models are
unavailable. The image and text encoders of $\phi_l$ are frozen. The ensemble thus provides robust supervision for
filtering unreliable text-image pairs.

Given $\mathcal{D}$ and $\phi_l$, for every text query $T_i$, we examine whether its corresponding
image $I_i$ appears within the top-$K$ retrieved images. Specifically, a text-to-image similarity matrix is computed over $\mathcal{D}$. The matrix is constructed using cosine similarity
between $\ell_2$-normalized global text and image feature embeddings,
$f_i^{t}$ and $f_j^{v}$, respectively. For $T_i$, similarity is computed against all images in $\mathcal{D}$:
\begin{equation}
\mathcal{S}_i =
\left[
S\!\left(f_i^{t}, f_1^{v}\right),
\; S\!\left(f_i^{t}, f_2^{v}\right),
\; \ldots,
\; S\!\left(f_i^{t}, f_{N}^{v}\right)
\right],
\end{equation}
where $S(\cdot,\cdot)$ denotes cosine similarity.

The similarity scores $\mathcal{S}_i$ are sorted in descending
order, inducing a ranking over all images. The retrieval rank for model $\phi_l$, denoted as
$\mathrm{rank}_{\phi_l}(T_i, I_i)$, is defined as the position of the ground-truth
image $I_i$ in this sorted list.

We define an indicator variable $\delta_i \in \{0,1\}$ to determine whether
a pair $(T_i, I_i)$ is valid under the ensemble of pretrained models.
A pair is considered valid if at least one pretrained model ranks the
corresponding image within its top-$K$ retrieval results:
\begin{equation}
\delta_i =
\begin{cases}
1, & \text{if } \exists\, l \in \{1,\dots,p\} \text{ such that }
\mathrm{rank}_{\phi_l}(T_i, I_i) \le K, \\
0, & \text{otherwise.}
\end{cases}
\end{equation}

Based on this criterion, we construct a curated noise-aware dataset
\begin{equation}
\mathcal{D}^{\text{clean}} =
\left\{ (T_i, I_i) \in \mathcal{D} \;\middle|\; \delta_i = 1 \right\}.
\end{equation}
which is subsequently used for training the unified model.

\subsection{Discriminative ID Learning framework (DIL)}

Prior methods introduce identity-specific supervision to enhance discriminative learning. While effective on individual datasets, such formulations do not scale well under unified training, where the number of identities grows significantly. The rapid expansion of the identity space leads to weak class separation with conventional classification objectives.

To address this, we design a more discriminative and scalable identity loss that enforces angular separation in a shared embedding space, enabling unified training under large-scale identity supervision. Let
the feature vectors $f^{v}$ and $f^{t}$ be obtained by the learnable image encoder $\phi_{img}$ and text encoder $\phi_{txt}$, respectively. 

To enhance inter-class separability and scale the number of unique identities, 
we introduce our Multimodal Angular Identity loss. We aim to increase the distance between unique IDs by an additive angular margin $m$.

Let $w_j \in \mathbb{R}^d$ denote the $\ell_2$-normalized
classification weight vector corresponding to the $j$-th identity class,
where $j \in \{1, \ldots, C\}$ and $C$ is the total number of classes. For a
given sample $i$, we define $\theta_{i,j}$ as the angle between $f_i^{v}$ and
$w_j$, given by
\begin{equation}
\cos \theta_{i,j} = f_i^{v} \cdot w_j .
\end{equation}
For the ground-truth class $y_i$, the corresponding angle is denoted as
$\theta_{i,y_i}$, which measures the angular similarity between the visual
feature $f_i^{v}$ and its true class weight $w_{y_i}$.

The margin-adjusted similarity for the target class is defined as
\begin{equation}
\gamma_{i,y_i} = \cos(\theta_{i,y_i} + m).
\end{equation}

To ensure monotonicity of the angular function and maintain numerical stability, the margin-adjusted similarity is defined in a piecewise manner as
\begin{equation}
\gamma_{i,y_i} =
\left\{
\begin{aligned}
&\cos(\theta_{i,y_i} + m),
&& \theta_{i,y_i} \le \pi - m, \\
&\cos(\theta_{i,y_i}) - m \sin(\pi - m),
&& \text{otherwise}.
\end{aligned}
\right.
\end{equation}

The final logits used for training are defined as
\begin{equation}
z_{i,j}^{v} =
\begin{cases}
s\gamma_{i,y_i}, & j = y_i, \\
s \gamma_{i,j}, & j \neq y_i,
\end{cases}
\end{equation}
where $s$ is a scaling factor that controls the magnitude of the logits.

The logits are then optimized using the standard cross-entropy loss:
\begin{equation}
\mathcal{L}_{img}
= -\frac{1}{N} \sum_{i=1}^{N}
\log
\frac{\exp(z_{i,y_i}^{v})}
{\sum_{j=1}^{C} \exp(z_{i,j}^{v})}.
\end{equation}

Similarly, text-based identity loss is computed using the textual embeddings $f^{t}$ and using the shared class weight $w$:
\begin{equation}
\mathcal{L}_{txt}
= -\frac{1}{N} \sum_{i=1}^{N}
\log
\frac{\exp(z_{i,y_i}^{t})}
{\sum_{j=1}^{C} \exp(z_{i,j}^{t})}.
\end{equation}

\textbf{Multimodal Angular Identity Loss.} The final identity supervision jointly optimizes both modalities,
\begin{equation}
\mathcal{L}_{\text{MA-ID}} = \frac{1}{2} \left( \mathcal{L}_{\text{img}} + \mathcal{L}_{\text{txt}} \right).
\end{equation}

This formulation enforces consistent angular discrimination across modalities, yielding compact intra-class clusters and well-separated inter-class boundaries, which is crucial for scalable unified TBPS training.
\subsection{Overall training objective}
We train the model using noise-aware curated dataset $\mathcal{D}^{\text{clean}}$. We adopt ranking loss TAL \cite{b16}. Our model ($\phi_{img}$ and $\phi_{txt}$) is optimized using the following overall objective, 
\begin{equation}
\mathcal{L} = \mathcal{L}_{\text{MA-ID}}+\mathcal{L}_{TAL}.
\end{equation}
\begin{table*}
\centering
\resizebox{\textwidth}{!}{
\begin{tabular}{|l|l|cccc|cccc|cccc|}
\hline
\multirow{2}{*}{\textbf{Methods}} 
& \multirow{2}{*}{\textbf{Venue}} 
& \multicolumn{4}{c|}{\textbf{CUHK-PEDES}} 
& \multicolumn{4}{c|}{\textbf{ICFG-PEDES}} 
& \multicolumn{4}{c|}{\textbf{RSTPReid}}  \\
\cline{3-14}
&  & R-1 & R-5 & R-10 & mAP 
& R-1 & R-5 & R-10 & mAP 
& R-1 & R-5 & R-10 & mAP  \\
\hline

\textit{CLIP based}&  &  &  &  &  
&  &  &  &  
&  &  &  &    \\
CFine \cite{b2}& TIP & 69.57 & 85.93 & 91.15 & - & 60.83 & 76.55 & 82.42 & - & 50.55 & 72.50 & 81.60 & -  \\ 
 CLIP~\cite{b12}& PMLR & 68.55 & 86.32 & 91.85 & 61.25 & 56.80 & 75.99 & 82.31 & 31.93 & 56.00 & 80.85 & 87.90 & 44.06  \\
IRRA \cite{b1}& CVPR & 73.38 & 89.93 & 93.71 & 66.13 & 63.46 & 80.25 & 85.82 & 38.06 & 60.20 & 81.30 & 88.20 & 47.17  \\
TBPS-CLIP \cite{b23} &AAAI& 73.54 &88.19& 92.35& 65.38&
65.05 &80.34& 85.47 &39.83 & 62.10& 81.90& 87.75& 48.00 \\

CFAM \cite{b41} & CVPR & 75.60& 90.53& 94.36& 67.27 &65.38& 81.17& 86.35& 39.42 &62.45& 83.55& 91.10& 49.50\\

 ICL \cite{b46}&CVPR  & 76.41& 90.48& 94.33& 68.04& 68.11& 82.59& 87.52&40.81 & 67.70 &86.05& 91.75& 52.62 \\
\hline
\textit{Non-CLIP based}&  &  &  &  &  
&  &  &  &  
&  &  &  &    \\
APTM~\cite{b22} & MM
& 76.53 & 90.04 & 94.15 & 66.91 
& {68.51} & 82.99 & 87.56 & 41.22 
& 67.50 & 85.70 & 91.45 & 52.56  \\
RaSa~\cite{b24} & IJCAI
& 76.51 & 90.29 & 94.25 & 69.38 
& 65.28 & 80.40 & 85.12 & 41.29 
& 66.90 & 86.50 & 91.35 & 52.31 \\
WoRA-TBPS~\cite{b39} & MM 
& 76.38 & 89.72 & 93.49 & 67.22 
& 68.35 & {83.10} & 87.53 & 42.60 
& 66.85 & 86.45 & 91.10 & 52.49 \\

\hline
RDE \cite{b16} & CVPR & 75.94 & 90.14 & 94.12 & 67.56 & 67.68 & 82.47 & 87.36 & 40.06 & 65.35 & 83.95 & 89.90 & 50.88  \\ 
RDE$^{*}$  & - & 76.32 & 90.95 & 94.43 & 68.13 & 66.73 & 82.87 & 87.63 & 39.42 & 65.30 & 85.90 & 90.90 & 51.63  \\
Scale-TBPS  & \textbf{Ours}
& 76.85 & 90.22 & 94.23 & 70.05
& 68.22 & 82.09 & 86.80 & 45.16
& 68.90 & 87.00 & 91.65 & 56.00 \\
Scale-TBPS$^{*}$ & \textbf{Ours}
& \textbf{77.91} & \textbf{91.07} & \textbf{94.49} & \textbf{70.88} 
& \textbf{68.24} & \textbf{83.00} & \textbf{87.63} & \textbf{45.68}
& \textbf{71.70} & \textbf{87.40} & \textbf{91.95} & \textbf{58.38} 
 \\
 \hline
\end{tabular}
}
\caption{Comparison study of Scale-TBPS with existing methods. `$^*$' denotes the results with NNN}
\label{comparison}
\end{table*}
\subsection{Test time Nearest-Neighbor-Normalization (NNN)}
We observe that a single text query may exhibit high similarity with multiple gallery images, resulting in biased retrieval scores. Motivated by this insight and inspired by Nearest Neighbor Normalization (NNN \cite{b50}), we apply similarity normalization after computing text-to-image similarities at test time. This lightweight post-processing strategy leads to a substantial improvement in retrieval performance without additional training.

Formally, the similarity between a query text $T$ and a reference image $I$ is defined as $ S(T, I) = f^t \cdot f^v$. For each retrieval candidate $I$, we define a bias term $b(I)$ as a scaled average similarity between $I$ and its top-$k$ nearest query texts. 
We denote the set of $\mathcal{K}$ queries from the reference query dataset $\mathcal{D}_{test}$ that achieve the highest similarity scores with $I$. The bias term is then computed as
\begin{equation}
b(I) = \alpha  \frac{1}{\mathcal{K}} \sum_{f^{t}_j \in \mathcal{D}_{\text{top-}\mathcal{K}}(I)} f^{t}_j \cdot f^v,
\end{equation}
where  $
\mathcal{D}_{\text{top-}\mathcal{K}}(I)$ denotes the set of top-$\mathcal{K}$ similar text candidates, retrieved from $\mathcal{D}_{\text{test}}$ with respect to the image $I$.

The normalized retrieval score is obtained by subtracting the estimated bias from the original similarity, $S_N(T, I) = S(T, I) - b(I)$.
In our experiments, we use both normalized and unnormalized similarity scores for computation of Rank-$k$ and mAP. Intuitively, NNN exploits the nearest $\mathcal{K}$ query embeddings to mitigate hubness and better differentiate visually similar retrieval candidates. When integrated with retrieval systems, this procedure introduces only sublinear overhead and adds a constant-factor cost to retrieval runtime. 

\section{Experiments}

In this section, we conduct extensive experiments to comprehensively evaluate the effectiveness and scalability of the proposed Scale-TBPS framework.

\subsection{Experimental Details}

For our unified dataset curation, we combine training set of four widely used text-based person search benchmarks: CUHK-PEDES \cite{b13}, ICFG-PEDES \cite{b6}, RSTPReid \cite{b5} and IIITD-20K \cite{b14}.  
We utilize three pretrained models trained individually on CUHK-PEDES, ICFG-PEDES, and RSTPReid to filter pairs. Notably, we intentionally exclude models trained on {IIITD-20K} in order to explicitly evaluate the scalability of our framework. Our unified dataset contains a total of 30,957 unique identities in the training set, compared to 11,003, 3,102, 3,701, 15000 training set identities in CUHK-PEDES, ICFG-PEDES, RSTPReid and IIITD-20K, respectively.

 We set $K$=25. Following \cite{b50}, we use $\alpha$ = 0.75 and $\mathcal{K}$ = 16, and do not tune them for any of the experiments. We employ a discriminative identity classifier with a fixed scale of $s=30$ and an angular margin of $m=0.35$. We use RDE as our baseline. We follow the training configurations of RDE.
We follow the standard test protocols used across all TBPS datasets.
We evaluate retrieval performance using Rank-$k$ accuracy at $k=\{1,5,10\}$, and mean Average Precision (mAP).

\subsection{Main results}

As shown in Table~\ref{comparison}, we compare Scale-TBPS with state-of-the-art dataset-centric methods, where each model is trained independently on a single dataset. In contrast, Scale-TBPS learns a single unified model and outperforms these dataset-specific approaches in most of the metrics. We report results for Scale-TBPS with and without test-time similarity normalization NNN, where Scale-TBPS$^*$ denotes the variant using normalization. Scale-TBPS uses CLIP as the backbone, but we report discussion with both CLIP-based and non-CLIP-based approaches. 

\textbf{Comparison against CLIP-based methods.}
Scale-TBPS consistently outperforms RDE across datasets and metrics. In particular, we see a significant gain in terms of mAP. We see a gain of approximately 5.1\% in mAP on both ICFG-PEDES and RSTPReid. 
In the transductive setting where test time normalization is applied, compared to RDE$^*$, Scale-TBPS$^*$ shows a 2.75\%, 6.26\%, and 6.75\% increase in mAP on CUHK-PEDES, ICFG-PEDES, and RSTPReid, respectively. On RSTPReid, our method shows a 6.40\% improvement in Rank-1 accuracy compared to RDE$^*$. 

The CLIP-based state-of-the-art method ICL \cite{b46} relies on multimodal large language models (MLLMs) \cite{b52} for data augmentation. Despite not employing such auxiliary components, Scale-TBPS surpasses ICL in both Rank-1 and mAP. Overall, our single unified model outperforms all CLIP-based methods across metrics.


To study the effect of NNN, we also report RDE$^*$. We use the values of $\alpha$ and $\mathcal{K}$ as suggested in \cite{b50} and do not tune them either for Scale-TBPS or RDE. RDE$^*$ shows only a nominal gain of 0.38\% in Rank-1 on CUHK-PEDES over RDE, whereas Scale-TBPS$^*$ shows a 1.15\% boost on the same metric compared to Scale-TBPS. Further, Rank-1 accuracy decreases for ICFG-PEDES and RSTPReid. While NNN effectively mitigates candidate bias $b(I)$ for Scale-TBPS, its benefit is less pronounced for RDE, where dataset-centric training calibrates similarity scores and reduces hubness.

\textbf{Comparison against non-CLIP-based methods.}
Scale-TBPS outperforms RaSa \cite{b24}, which is ALBEF-based, even though ALBEF is generally considered a stronger backbone than CLIP. On ICFG-PEDES it shows a 2.94\% gain in Rank-1, while on RSTPReid it shows a 2\% gain. We also compare against APTM and WoRA-TBPS, whose backbones are pre-trained on pedestrian-specific data and thus exhibit strong capabilities. Notably, CLIP-based methods are significantly lighter in terms of parameter count. Compared to APTM \cite{b48} and WoRA-TBPS, our method outperforms APTM on both CUHK-PEDES and RSTPReid. In contrast, the baseline RDE fails to surpass APTM and RaSa, whereas our method achieves consistent improvements over both. 
These results demonstrate that properly designed unified model can outperform different dataset-centric models. Our CLIP based method gives competing performance against stronger backbones too.

\textbf{Results on IIITD-20K.}
We do not use IIITD-20K trained model to filter data. We deliberately exclude model trained on IIITD-20K to explicitly evaluate scalability of our framework. We find that our method with and without test time normalization outperforms the baseline RDE by a significant margin. We observe 1.52\% increase in Rank-1 accuracy in Scale-TBPS and an additional 2.78\% accuracy increase are observed with Scale-TBPS$^{*}$ as reported in Table \ref{tab:IIITD_comparison}. Thus the substantial increase validates the scalability of our method.
\begin{table}
\centering

\begin{tabular}{l|cccc}
\hline
Method & R-1 & R-5 & R-10 & mAP  \\
\hline
CLIP  & 73.48 & 91.82 & 95.60 & 81.69  \\
IRRA~\cite{b1}  & 77.08 & --    & --    & --    \\
RDE~\cite{b16}  & 80.90 & 94.48 & 97.02 & 87.04  \\
Scale-TBPS (Ours)& 82.42 & 95.50    & 97.42   & 88.28    \\
Scale-TBPS$^{*}$ (Ours)  & \textbf{85.20} & \textbf{96.56}    & \textbf{98.04}  & \textbf{90.31}   \\ 
\hline
\end{tabular}
\caption{Comparison results on IIITD-20K}
\label{tab:IIITD_comparison}
\end{table}

\textbf{Results on UFine6926.}
As UFine6926 \cite{b41} is not included in our training paradigm, we evaluate our model in a zero-shot setting on this dataset, as reported in Table~\ref{tab:ufinebench}. For comparison, we report results from prior methods that are trained exclusively on CUHK-PEDES, as this setting yields the strongest performance in their respective works. Our Scale-TBPS shows a 4.5\% gain in Rank-1 and a 4.06\% gain in mAP. It is observed that Scale-TBPS$^*$ provides a significant boost in zero-shot performance, notably a 9.23\% gain in Rank-1 compared to Scale-TBPS. Hence, our proposed method generalizes well to unseen domain.
\begin{table}
\centering

\begin{tabular}{lcccc}
\hline
Method & R-1 &R-5&R-10& mAP  \\
\hline
IRRA~\cite{b1} & 37.51 & 54.92 & 64.29 &40.76 \\
RDE~\cite{b16} & 40.37 & 57.49 & 66.05 & 42.68\\
ICL~\cite{b46}& 46.40& 63.55 & 72.08 & 48.68  \\
Scale-TBPS (Ours) & 44.87 & 63.57 & 72.13& 46.74 \\
Scale-TBPS$^*$ (Ours) & \textbf{54.10} & \textbf{71.76} & \textbf{79.26}& \textbf{56.00}  \\
\hline
\end{tabular}
\caption{Zero-shot Performance Comparison on UFine6926}
\label{tab:ufinebench}
\end{table}
\subsection{Ablation and Hyperparameter Study}

We analyze the contribution of the two core components of Scale-TBPS in Table \ref{tab:ablation}:  
(i) {noise-aware unified data curation} (NDC), and  
(ii) {discriminative identity learning framework} (DIL) for scaling to a large number of unique identities.

For fair comparison, we include two strong naive joint training baselines, where IRRA and RDE are individually trained with the joint dataset. We keep RDE's original filtering protocols. As reported in Table \ref{tab:ablation}, even though naive joint training leverages a larger volume of training data, it fails to yield better results across all datasets. In particular, it underperforms compared to RDE model on ICFG-PEDES. We also report the performance of our modules NDC and DIL independently.

Compared to naive joint training, NDC improves both Rank-1 and mAP on ICFG-PEDES and RSTPReid. On CUHK-PEDES, NDC shows a marginal drop. On the other hand, DIL shows significant improvement over the naive method on ICFG-PEDES. Our proposed method NDC combined with DIL outperforms naive method by 1\% in CUHK-PEDES and nearly 2\% in ICFG-PEDES and RSTPReid in Rank-1 accuracy. In terms of mAP, our method outperforms the naive approach by 1.73\% on CUHK-PEDES, 4.48\% on ICFG-PEDES, and 1.86\% on RSTPReid.
\begin{table}[t]
\centering

\resizebox{\columnwidth}{!}{
\begin{tabular}{l|cc|cc|cc}
\hline
Method 
& \multicolumn{2}{c|}{CUHK-PEDES} 
& \multicolumn{2}{c|}{ICFG-PEDES} 
& \multicolumn{2}{c}{RSTPReid} \\
& R-1 & mAP 
& R-1 & mAP 
& R-1 & mAP \\
\hline
Naive-IRRA
& 73.26 & 66.70
& 62.34 & 38.98
& 65.05& 53.22 \\
Naive-RDE
& 76.40 & 68.32 
& 66.03 & 40.68 
& 67.00 & 54.14 \\

NDC 
& 75.97 & 68.02 
& 66.42 & 40.92 
& 70.15 & 54.75 \\

DIL
& 76.09 & 68.96 
& 66.60 & 42.81 
& 66.60 & 54.30 \\

\textbf{NDC+DIL}
& \textbf{76.85} & \textbf{70.05}
& \textbf{68.22} & \textbf{45.16}
& \textbf{68.90} & \textbf{56.00} \\
\hline
\end{tabular}
}
\caption{Ablation study for NDC and DIL. NDC+DIL is the Scale-TBPS method}
\label{tab:ablation}
\end{table}

\begin{figure}[t]
    \centering
    \includegraphics[width=\columnwidth,keepaspectratio]{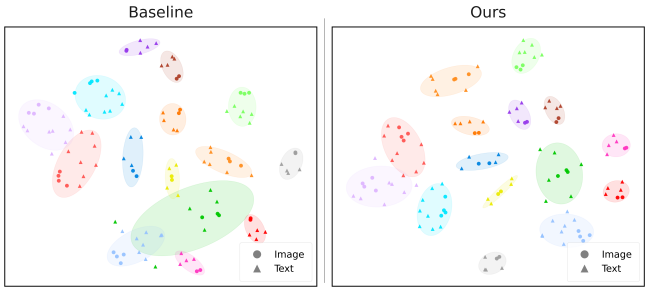}
    \caption{t-SNE visualization comparing naive joint training and the proposed DIL. Different color signifies different IDs. Our method exhibits tighter intra-class clustering and improved inter-class separation}
    \label{fig:single_column}
\end{figure}

\textbf{Analysis of DIL.}
 To demonstrate the effectiveness of the proposed DIL framework in separating identity classes, we visualize the learned feature distributions using t-SNE \cite{b59}. We compare our method against naive joint training. As illustrated in Figure~\ref{fig:single_column}, the proposed method yields well-structured clusters characterized by strong intra-class compactness and clear inter-class separability. In contrast, naive joint training produces overlapping clusters, indicating weaker identity discrimination. These observations confirm that DIL effectively enhances discriminative feature learning under a large number of identities. 

\begin{figure}[t]
    \centering
\includegraphics[width=.9\columnwidth,keepaspectratio]{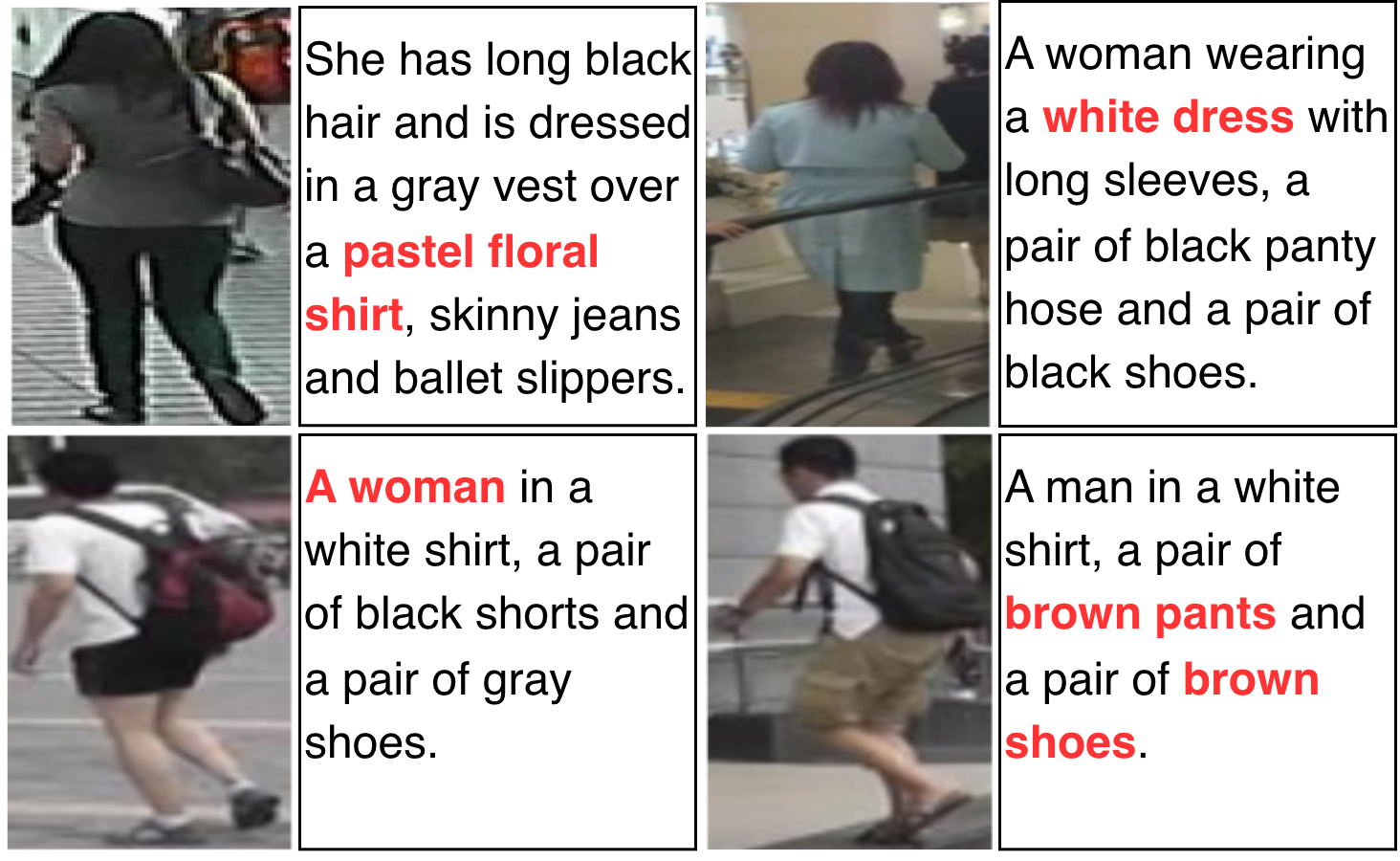}
    \caption{Examples of noisy samples removed by our NDC module. Noisy parts are highlighted in red}
    \label{fig:noisy}
\end{figure}

\textbf{Analysis of NDC.} 
We show qualitative examples that our NDC module filters out in Figure \ref{fig:noisy}. Since pair selection is guided by  three diverse TBPS models, the filtering procedure is able to remove noisy pairs while retaining a substantial portion of useful data. NDC module is a one-time preprocessing step that takes only 556.47 seconds on a single Nvidia A100 GPU to filter the entire joint dataset of 1,69,810 text-image pairs, making it easily scalable to large volumes of data.

\begin{figure}
    \centering
    \includegraphics[width=.9\columnwidth,keepaspectratio]{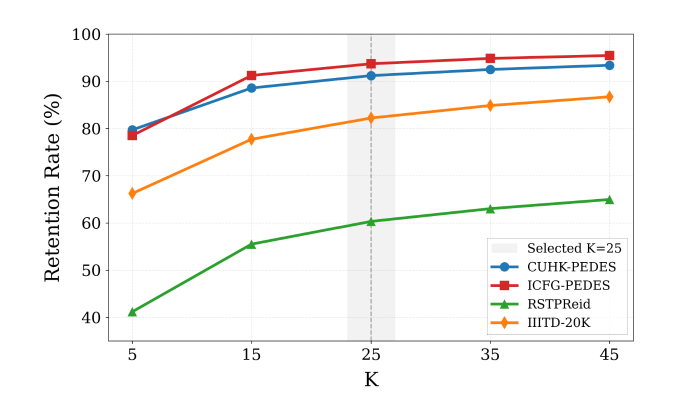}
    \caption{Data retention rates (\%) after filtration by our NDC module across various values of $K$ for different TBPS datasets.  }
    
    \label{fig:dataset}
\end{figure}

In Figure ~\ref{fig:dataset} we see for $K$=25, CUHK-PEDES and ICFG-PEDES retains approximately 90\% of their data. In contrast, RSTPReid retains only 60.3\% of the data. This is consistent with prior work, which identifies RSTPReid as one of the noisiest TBPS benchmarks. We intentionally exclude any model trained on IIITD-20K during the filtering stage. Even under this constraint, approximately 82\% of the IIITD-20K data is retained, validating that our method can effectively scale to new TBPS distributions using a limited set of pretrained TBPS experts.

\textbf{Hyperparameter Analysis.}
We conduct a hyperparameter sensitivity analysis on the margin ($m$) and scale ($s$) used in our DIL. $m$ is used to set the boundary between different identity classes, and $s$ signifies the sharpness of logits in DIL. The hyperparameters are selected based on Rank-1 accuracy on CUHK-PEDES following previous TBPS methods. 

\begin{figure}
    \centering
\includegraphics[width=.9\columnwidth]{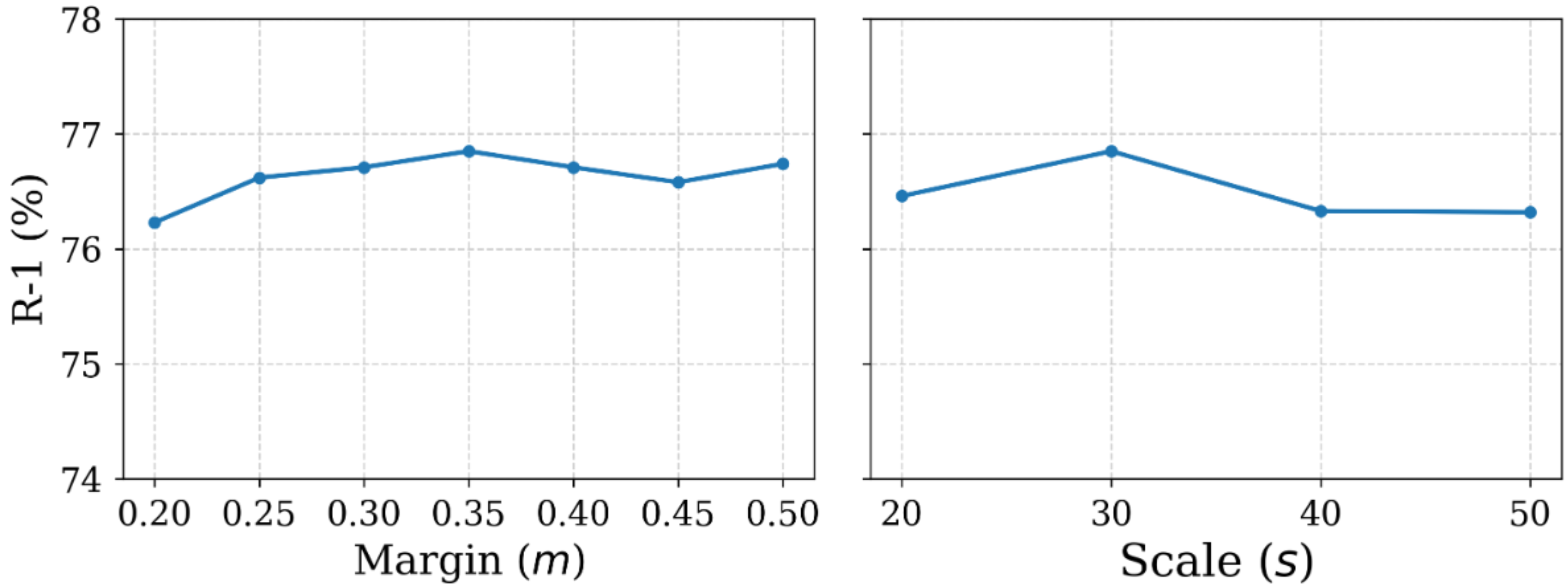}
\caption{Hyperparameter Sensitivity Analysis on CUHK-PEDES}
\label{fig:hyperparam}
\end{figure}

In Figure \ref{fig:hyperparam} we show that the best results are achieved with $m$ = 0.35 and $s$ = 30. The proposed method exhibits strong robustness to hyperparameter variations, with Rank-1 accuracy fluctuating by less than 1\% across a wide range of values. 

 In Figure~\ref{fig:dataset}, we present a comprehensive study of dataset percentage retention with respect to different $K$ values. Our NDC module selects a pair when the corresponding image is retrieved within the Top-$K$ samples for a given caption. We observe that at $K = 25$, the retention curve plateaus, and we choose this value for our experiments, as below this point the data retention drops, while beyond it the increase is only marginal. 
 
 We provide code and additional results in the supplementary material.

\section{Conclusion}
In this work, we propose Scale-TBPS, a unified training paradigm that enables learning a single model across multiple text-based person search datasets. We introduce a noise-aware data curation strategy to cohesively merge multiple datasets, along with a discriminative ID learning framework designed to effectively handle the rapid growth in unique identities that arises with dataset merging. Comprehensive experiments on CUHK-PEDES, RSTPReid, ICFG-PEDES, and IIITD-20K demonstrate that a single Scale-TBPS model consistently matches or surpasses dataset-specific state-of-the-art methods, as well as naive joint training, under diverse distribution shifts. Furthermore, we demonstrate zero-shot robustness on UFine6926, highlighting strong generalization to previously unseen data. Additionally, we observe that test time normalization significantly boosts the performance of Scale-TBPS compared to other methods.

\bibliographystyle{named}
\bibliography{ijcai26}

@inproceedings{b1,
  title={Cross-modal implicit relation reasoning and aligning for text-to-image person retrieval},
  author={Jiang, Mang},
  booktitle={CVPR},
  pages={2787--2797},
  year={2023}
}

@article{b2,
  title={Clip-driven fine-grained text-image person re-identification},
  author={Yan, Shuanglin and Dong, Neng and Zhang, Liyan and Tang, Jinhui},
  journal={TIP},
  volume={32},
  pages={6032--6046},
  year={2023},
  publisher={IEEE}
}

@inproceedings{b5,
  title={Dssl: Deep surroundings-person separation learning for text-based person retrieval},
  author={Zhu, Aichun and Wang, Zijie and Li, Yifeng and Wan, Xili and Jin, Jing and Wang, Tian and Hu, Fangqiang and Hua, Gang},
  booktitle={ACM MM},
  pages={209--217},
  year={2021}
}

@article{b6,
  title={Semantically self-aligned network for text-to-image part-aware person re-identification},
  author={Ding, Zefeng and Ding, Changxing and Shao, Zhiyin and Tao, Dacheng},
  journal={arXiv preprint arXiv:2107.12666},
  year={2021}
}

@article{b8,
  title={TIPCB: A simple but effective part-based convolutional baseline for text-based person search},
  author={Chen, Yuhao and Zhang, Guoqing and Lu, Yujiang and Wang, Zhenxing and Zheng, Yuhui},
  journal={Neurocomputing},
  volume={494},
  pages={171--181},
  year={2022},
  publisher={Elsevier}
}

@inproceedings{b10,
  title={Learning granularity-unified representations for text-to-image person re-identification},
  author={Shao, Zhiyin and Zhang, Xinyu and Fang, Meng and Lin, Zhifeng and Wang, Jian and Ding, Changxing},
  booktitle={ACM MM},
  pages={5566--5574},
  year={2022}
}

@inproceedings{b12,
  title={Learning transferable visual models from natural language supervision},
  author={Radford, Alec and Kim, Jong Wook and Hallacy, Chris and Ramesh, Aditya and Goh, Gabriel and Agarwal, Sandhini and Sastry, Girish and Askell, Amanda and Mishkin, Pamela and Clark, Jack and others},
  booktitle={ICML},
  pages={8748--8763},
  year={2021},
  organization={PmLR}
}

@inproceedings{b13,
  title={Person search with natural language description},
  author={Li, Shuang and Xiao, Tong and Li, Hongsheng and Zhou, Bolei and Yue, Dayu and Wang, Xiaogang},
  booktitle={CVPR},
  pages={1970--1979},
  year={2017}
}

@inproceedings{b14,
  title={Dense captioning for Text-Image ReID},
  author={Subramanyam, A Venkata and Dubey, Vibhu and Sundararajan, Niranjan and Lall, Brejesh},
  booktitle={ICVGIP},
  pages={1--8},
  year={2023}
}

@inproceedings{b16,
  title={Noisy-correspondence learning for text-to-image person re-identification},
  author={Qin, Yang and Chen, Yingke and Peng, Dezhong and Peng, Xi and Zhou, Joey Tianyi and Hu, Peng},
  booktitle={CVPR},
  pages={27197--27206},
  year={2024}
}

@inproceedings{b21,
  title={Align before fuse: Vision and language representation learning with momentum distillation},
  author={Li, Junnan and Selvaraju, Ramprasaath and Gotmare, Akhilesh and Joty, Shafiq and Xiong, Caiming and Hoi, Steven Chu Hong},
  booktitle={NeurIPS},
  volume={34},
  pages={9694--9705},
  year={2021}
}

@inproceedings{b22,
  title={Towards unified text-based person retrieval: A large-scale multi-attribute and language search benchmark},
  author={Yang, Shuyu and Zhou, Yinan and Zheng, Zhedong and Wang, Yaxiong and Zhu, Li and Wu, Yujiao},
  booktitle={ACM MM},
  pages={4492--4501},
  year={2023}
}

@inproceedings{b23,
  title={An empirical study of clip for text-based person search},
  author={Cao, Min and Bai, Yang and Zeng, Ziyin and Ye, Mang and Zhang, Min},
  booktitle={AAAI},
  volume={38},
  pages={465--473},
  year={2024}
}

@inproceedings{b24,
  title     = {RaSa: Relation and Sensitivity Aware Representation Learning for Text-based Person Search},
  author    = {Bai, Yang and Cao, Min and Gao, Daming and Cao, Ziqiang and Chen, Chen and Fan, Zhenfeng and Nie, Liqiang and Zhang, Min},
  booktitle = {Proceedings of the Thirty-Second International Joint Conference on
               Artificial Intelligence, {IJCAI-23}},
  publisher = {International Joint Conferences on Artificial Intelligence Organization},
  editor    = {Edith Elkind},
  pages     = {555--563},
  year      = {2023},
  month     = {8},
  note      = {Main Track},
  doi       = {10.24963/ijcai.2023/62},
  url       = {https://doi.org/10.24963/ijcai.2023/62},
}

@inproceedings{b38,
  title={Does prior data matter? Exploring joint training in the context of few-shot class-incremental learning},
  author={Kim, Shiwon and Hwang, Dongjun and Woo, Sungwon and Singh, Rita},
  booktitle={Proceedings of the IEEE/CVF International Conference on Computer Vision},
  pages={5185--5194},
  year={2025}
}

@inproceedings{b39,
  title={From data deluge to data curation: A filtering-wora paradigm for efficient text-based person search},
  author={Sun, Jintao and Fei, Hao and Ding, Gangyi and Zheng, Zhedong},
  booktitle={Proceedings of the ACM on Web Conference 2025},
  pages={2341--2351},
  year={2025}
}

@inproceedings{b40,
  title={Arcface: Additive angular margin loss for deep face recognition},
  author={Deng, Jiankang and Guo, Jia and Xue, Niannan and Zafeiriou, Stefanos},
  booktitle={Proceedings of the IEEE/CVF conference on computer vision and pattern recognition},
  pages={4690--4699},
  year={2019}
}

@inproceedings{b41,
  title={Ufinebench: Towards text-based person retrieval with ultra-fine granularity},
  author={Zuo, Jialong and Zhou, Hanyu and Nie, Ying and Zhang, Feng and Guo, Tianyu and Sang, Nong and Wang, Yunhe and Gao, Changxin},
  booktitle={Proceedings of the IEEE/CVF Conference on Computer Vision and Pattern Recognition},
  pages={22010--22019},
  year={2024}
}

@inproceedings{b42,
  title={Conceptual 12m: Pushing web-scale image-text pre-training to recognize long-tail visual concepts},
  author={Changpinyo, Soravit and Sharma, Piyush and Ding, Nan and Soricut, Radu},
  booktitle={Proceedings of the IEEE/CVF conference on computer vision and pattern recognition},
  pages={3558--3568},
  year={2021}
}

@article{b43,
  title={Yfcc100m: The new data in multimedia research},
  author={Thomee, Bart and Shamma, David A and Friedland, Gerald and Elizalde, Benjamin and Ni, Karl and Poland, Douglas and Borth, Damian and Li, Li-Jia},
  journal={Communications of the ACM},
  volume={59},
  number={2},
  pages={64--73},
  year={2016},
  publisher={ACM New York, NY, USA}
}

@inproceedings{b44,
  title={Unihcp: A unified model for human-centric perceptions},
  author={Ci, Yuanzheng and Wang, Yizhou and Chen, Meilin and Tang, Shixiang and Bai, Lei and Zhu, Feng and Zhao, Rui and Yu, Fengwei and Qi, Donglian and Ouyang, Wanli},
  booktitle={Proceedings of the IEEE/CVF conference on computer vision and pattern recognition},
  pages={17840--17852},
  year={2023}
}

@article{b45,
  title={MSCI: Addressing CLIP's Inherent Limitations for Compositional Zero-Shot Learning},
  author={Wang, Yue and Xu, Shuai and Zhu, Xuelin and Li, Yicong},
  journal={arXiv preprint arXiv:2505.10289},
  year={2025}
}

@inproceedings{b46,
  title={Human-centered Interactive Learning via MLLMs for Text-to-Image Person Re-identification},
  author={Qin, Yang and Chen, Chao and Fu, Zhihang and Peng, Dezhong and Peng, Xi and Hu, Peng},
  booktitle={Proceedings of the Computer Vision and Pattern Recognition Conference},
  pages={14390--14399},
  year={2025}
}

@inproceedings{b47,
  title     = {Towards Anytime Retrieval: A Benchmark for Anytime Person Re-Identification},
  author    = {Li, Xulin and Lu, Yan and Liu, Bin and Li, Jiaze and Yang, Qinhong and Gong, Tao and Chu, Qi and Ye, Mang and Yu, Nenghai},
  booktitle = {Proceedings of the Thirty-Fourth International Joint Conference on
               Artificial Intelligence, {IJCAI-25}},
  publisher = {International Joint Conferences on Artificial Intelligence Organization},
  editor    = {James Kwok},
  pages     = {1467--1475},
  year      = {2025},
  month     = {8},
  note      = {Main Track},
  doi       = {10.24963/ijcai.2025/164},
  url       = {https://doi.org/10.24963/ijcai.2025/164},
}

@inproceedings{b48,
  title={Towards unified text-based person retrieval: A large-scale multi-attribute and language search benchmark},
  author={Yang, Shuyu and Zhou, Yinan and Zheng, Zhedong and Wang, Yaxiong and Zhu, Li and Wu, Yujiao},
  booktitle={Proceedings of the 31st ACM international conference on multimedia},
  pages={4492--4501},
  year={2023}
}

@inproceedings{b50,
    title = "Nearest Neighbor Normalization Improves Multimodal Retrieval",
    author = "Chowdhury, Neil  and
      Wang, Franklin  and
      Shenoy, Sumedh  and
      Kiela, Douwe  and
      Schwettmann, Sarah  and
      Thrush, Tristan",
    editor = "Al-Onaizan, Yaser  and
      Bansal, Mohit  and
      Chen, Yun-Nung",
    booktitle = "Proceedings of the 2024 Conference on Empirical Methods in Natural Language Processing",
    month = nov,
    year = "2024",
    address = "Miami, Florida, USA",
    publisher = "Association for Computational Linguistics",
    url = "https://aclanthology.org/2024.emnlp-main.1257/",
    doi = "10.18653/v1/2024.emnlp-main.1257",
    pages = "22571--22582"
}

@inproceedings{b51,
  title={Harnessing the power of mllms for transferable text-to-image person reid},
  author={Tan, Wentan and Ding, Changxing and Jiang, Jiayu and Wang, Fei and Zhan, Yibing and Tao, Dapeng},
  booktitle={Proceedings of the IEEE/CVF Conference on Computer Vision and Pattern Recognition},
  pages={17127--17137},
  year={2024}
}

@article{b52,
  title={Qwen2-vl: Enhancing vision-language model's perception of the world at any resolution},
  author={Wang, Peng and Bai, Shuai and Tan, Sinan and Wang, Shijie and Fan, Zhihao and Bai, Jinze and Chen, Keqin and Liu, Xuejing and Wang, Jialin and Ge, Wenbin and others},
  journal={arXiv preprint arXiv:2409.12191},
  year={2024}
}

@inproceedings{b53,
  title={Mnemonics training: Multi-class incremental learning without forgetting},
  author={Liu, Yaoyao and Su, Yuting and Liu, An-An and Schiele, Bernt and Sun, Qianru},
  booktitle={Proceedings of the IEEE/CVF conference on Computer Vision and Pattern Recognition},
  pages={12245--12254},
  year={2020}
}

@article{b54,
  title={Deep learning face representation by joint identification-verification},
  author={Sun, Yi and Chen, Yuheng and Wang, Xiaogang and Tang, Xiaoou},
  journal={Advances in neural information processing systems},
  volume={27},
  year={2014}
}

@inproceedings{b57,
  title={Mixture-of-scores: Robust image-text data valuation via three lines of code},
  author={Wu, Sitong and Tan, Haoru and Chen, Yukang and Zhang, Shaofeng and Li, Jingyao and Yu, Bei and Qi, Xiaojuan and Jia, Jiaya},
  booktitle={Proceedings of the IEEE/CVF International Conference on Computer Vision},
  pages={24603--24614},
  year={2025}
}

@article{b59,
  title={Visualizing data using t-SNE},
  author={Maaten, Laurens van der and Hinton, Geoffrey},
  journal={Journal of machine learning research},
  volume={9},
  number={Nov},
  pages={2579--2605},
  year={2008}
}

\end{document}